\documentclass[runningheads]{llncs}
\usepackage{graphicx}
\usepackage{amsmath}
\usepackage{amssymb}
\usepackage{booktabs}
\usepackage{multirow}
\usepackage{pifont}
\usepackage{nicefrac}
\usepackage{comment}
\usepackage{wrapfig}
\usepackage{subfig}
\usepackage{cite}
\newcommand{\cmark}{\ding{51}}%
\newcommand{\xmark}{\ding{55}}%
\usepackage[pagebackref,breaklinks,colorlinks]{hyperref}

\mathchardef\mhyphen="2D

\usepackage{color}
\usepackage{xcolor}
\usepackage{caption}
\usepackage{mmstyle}
\usepackage[capitalize]{cleveref}
\crefname{section}{Sec.}{Secs.}
\Crefname{section}{Section}{Sections}
\Crefname{table}{Table}{Tables}
\crefname{table}{Tab.}{Tabs.}

\usepackage[accsupp]{axessibility}  

\usepackage{times}
\usepackage{epsfig}
\usepackage{tabulary}
\usepackage{url}

\newcommand{\red}[1]{{\color{red}#1}}

\newcommand{\blue}[1]{{\color{blue}#1}}
\newcommand{\inc}[1]{\blue{$\uparrow #1$}}
\newcommand{\rinc}[1]{\red{$\uparrow #1$}}
\newcommand{\dec}[1]{\red{$\downarrow #1$}}
\newcommand{\bdec}[1]{\blue{$\downarrow #1$}}
\def\x{$\times$}

\newcommand{\tablestyle}[2]{\setlength{\tabcolsep}{#1}\renewcommand{\arraystretch}{#2}\centering\footnotesize}
\newlength\savewidth\newcommand\shline{\noalign{\global\savewidth\arrayrulewidth
		\global\arrayrulewidth 1pt}\hline\noalign{\global\arrayrulewidth\savewidth}}
\begin{document}
\pagestyle{headings}
\mainmatter

\title{Mitigating Representation Bias in Action Recognition: \\
		Algorithms and Benchmarks}
\titlerunning{Mitigating Representation Bias in Action Recognition: Algorithms and Benchmarks} 
\authorrunning{Haodong Duan, Yue Zhao, Kai Chen, Yuanjun Xiong, Dahua Lin} 

\author{Haodong Duan\inst{1, 3} \and
Yue Zhao\inst{2} \and Kai Chen\inst{3} \and Yuanjun Xiong\inst{4} \and
Dahua Lin\inst{1}}

\institute{The Chinese University of Hong Kong \and The University of Texas at Austin \and Shanghai AI Lab \and Amazon AI}

\maketitle

\begin{abstract}
Deep learning models have achieved excellent recognition results on large-scale video benchmarks.
However, they perform poorly when applied to videos with rare scenes or objects, primarily due to the bias of existing video datasets.
We tackle this problem from two different angles: algorithm and dataset.
From the perspective of algorithms, we propose Spatial-aware Multi-Aspect Debiasing (\textbf{SMAD}), 
which incorporates both \textit{explicit} debiasing with multi-aspect adversarial training 
and \textit{implicit} debiasing with the spatial actionness reweighting module,
to learn a more generic representation invariant to non-action aspects.
To neutralize the intrinsic dataset bias, 
we propose \textbf{OmniDebias} to leverage web data for joint training selectively, 
which can achieve higher performance with far fewer web data.
To verify the effectiveness, 
we establish evaluation protocols and perform extensive experiments on both re-distributed splits of existing datasets 
and a new evaluation dataset focusing on the action with rare scenes.
We also show that the debiased representation can generalize better when transferred to other datasets and tasks.

\end{abstract}
\section{Introduction}

Human beings have cognitive bias, and so do the machine learning systems~\cite{mehrabi2019survey}.
Human cognitive bias comes from the uniqueness of individual experiences (learning materials)
and the tendency of brains to simplify information processing~\cite{haselton2015evolution}.
Machine learning systems are biased for similar reasons.
First, the datasets used for training can be intrinsically biased:
\eg, sampled from a shifted distribution~\cite{torralba2011unbiased}
or collected with a pre-defined ontology~\cite{ray2018scenes}.
Even if the dataset faithfully represents the real world, 
there is human bias in the real world which we do not want the machine learning system to exploit, 
\eg, gender bias~\cite{caliskan2017semantics, bolukbasi2016man}.
Mitigating the bias in machine learning systems has long been 
a challenging yet valuable research area~\cite{barocas2016big, dwork2012fairness, feldman2015certifying}.

In computer vision, the efforts for building datasets that faithfully represent the real visual world never end.
Better data collection and labeling strategies~\cite{deng2009imagenet, zhou2017places, thomee2016yfcc100m, ray2018scenes} 
are designed for building less biased datasets from scratch.
Besides, various tools can be applied to a built dataset 
(visual~\cite{wang2020revise} or tabular~\cite{bellamy2018ai} data) 
to detect and mitigate unwanted bias.
In action recognition, \cite{li2018resound} introduces the concept of representation bias and attempts to 
reduce it throughout dataset construction.
However, the dataset they propose is on a small scale and in a narrow domain.
We investigate existing large-scale datasets instead and 
quantify the representation bias by designing different 
train-test splits and analyzing the performance gaps.

Besides, we propose \textbf{OmniDebias}, which uses external web media as auxiliary data to mitigate the dataset bias.
On the one hand, the diversity of web data provide us with rich examples that are uncommon in existing datasets, 
which makes it a suitable data source for debiasing.
On the other hand, web data are also severely biased to some factors, \eg, \emph{scene}.
OmniDebias adopts a simple yet effective data selection strategy
to sample a less biased subset from the entire dataset.
Co-training with the selected subset outperforms the vanilla co-training both in performance and data efficiency.

Though effective in debiasing, 
constructing `unbiased' datasets can be difficult and may cost lots of human labor,
while designing debiasing algorithms is a much cheaper alternative.
A series of works~\cite{zhang2018mitigating, madras2018learning} aim at devising algorithms to mitigate the bias in the learned representation, 
preventing the algorithms from amplifying the bias in training data.
In particular, SDN~\cite{choi2019can} proposes to mitigate \emph{scene} bias in action recognition 
with adversarial training and human mask confusion loss.
Previous works usually restrict the debiasing algorithm to a specific factor.
In the real world, 
the bias in the dataset can be complex and non-trivial to understand.
To deal with more complicated dataset bias, 
we extend the single-factor adversarial training to a multi-aspect fashion, 
which shows better generalization capability.

To mitigate the \emph{generic} representation bias in action recognition,
we propose a spatial-aware multi-aspect debiasing framework (\textbf{SMAD}).
A video can have multiple facets besides the action label,
such as the background \emph{scene} or the \emph{object} that people interact with.
Video datasets collected for different purposes may emphasize different facets.
Considering this characteristic, 
we propose multi-aspect adversarial training (\textbf{MAAT}) to enforce the model invariant to these \emph{non-action} facets.
We also introduce Spatial-Aware Actionness Reweighting (\textbf{SAAR}) to ensure that the model learns where to focus 
to recognize action without being affected by features related to other facets.
The framework \textbf{SMAD} proves to be generic for videos with various kinds of bias and does not depend on extra knowledge of specific datasets.

To fairly exhibit the effectiveness of the proposed debiasing algorithm, 
we devise a series of evaluation protocols.
First, for the existing large-scale dataset Kinetics-400~\cite{carreira2017quo},
we re-distribute the original train splits by either \emph{scene} or \emph{object} such that the hidden facet does not overlap between 
the \emph{re-distributed} train and test sets (\textbf{facet-based re-distribution}).
Second, we collect an additional Action with RAre Scene (ARAS) dataset\footnote{
    Dataset released at \url{https://github.com/kennymckormick/ARAS-Dataset}. }
for evaluation to simulate the \textbf{out-of-distribution} setting.
Third, we follow the routine of measuring the debiasing effect by transferring the learned model to downstream tasks 
(\textbf{downstream-task transferring}), such as feature classification, few-shot learning, 
and finetuning on other datasets (as is proposed by \cite{choi2019can}). 

Our contributions are three-fold:
1. We propose SMAD, which considers multiple aspects in adversarial training and 
achieves better performance when complex bias exists in the training set.
2. We propose OmniDebias, which exploits the richness and diversity of web data effectively and efficiently. 
3. We evaluate our method on both conventional evaluation protocols (downstream-task transferring) 
as well as the new ones (facet-based re-distribution, out-of-distribution testing). 
The improvements of our methods on all three benchmarks are consistent and remarkable. 

\section{Related Work}

\textbf{Action Recognition.}
Action recognition aims at recognizing human activities in videos.
Following the success of deep learning in the image domain, 
two series of deep ConvNets become the mainstream architectures for action recognition, 
named 2D-CNN and 3D-CNN methods.
2D-CNN methods like Two-Stream~\cite{simonyan2014two} and TSN~\cite{wang2016temporal} are light-weight 
while lacking temporal modeling capability to some extent. 
3D-CNN methods~\cite{carreira2017quo, feichtenhofer2019slowfast, tran2018closer, tran2019video} use 3D convolutions for temporal modeling 
and achieve the state-of-the-art on large-scale benchmarks like Kinetics-400~\cite{carreira2017quo}.
In this paper, we show that both architectures are vulnerable to biases.
Our proposed framework can help to mitigate this problem.

\textbf{Mitigating Dataset Bias.}
All datasets, more or less, have dataset bias.
In computer vision, \cite{torralba2011unbiased} studies 12 widely used image datasets and 
finds their data are of different domains and distant from the real visual world. 
In natural language processing, gender bias occurs in corpus collected from social media and news~\cite{jia2016women,garcia2014gender}.
There are two main approaches to mitigate dataset bias:
The first is to design better data collection and labeling strategies~\cite{ray2018scenes}
or to calibrate the existing dataset with bias detection tools~\cite{wang2020revise, holland2018dataset}.
The second is to compensate dataset bias with domain adaptation techniques~\cite{khosla2012undoing, fernando2013unsupervised, oquab2014learning}.
In this paper, following the first approach, 
we propose to use diversified web media to neutralize the dataset bias.

\textbf{Mitigating Algorithm Bias. }
Even if the dataset faithfully represents the real world, bias still exists.
Due to human bias, real-world data may bias towards specific factors, 
while discriminative models even amplify such unwanted bias~\cite{zhao2017men}.
In machine learning, it is intuitive to add constraints or regularizations for the pursued fairness metric 
to the existing optimization objective~\cite{woodworth2017learning, agarwal2018reductions, zafar2017fairness}.
However, most of these approaches are intractable in deep learning.
Meanwhile, adversarial training has broader applications both in machine learning and deep learning.
\cite{zhang2018mitigating} use adversarial debiasing for bias mitigation,
but the bias factor is required to be known beforehand.
\cite{li2019repair} propose to use adversarial example reweighting and 
achieves good performance on debiasing action recognition.

\textbf{Domain adaptation. }
Domain adaptation (DA) aims at learning well-performing models 
on the target domain with training data from the source domain.
To that end, many works try to find a common feature space for the source and target domains
via adversarial training, both for image tasks~\cite{ganin2015unsupervised, saito2018maximum, chen2018domain}
and action recognition~\cite{choi2020shuffle,choi2020unsupervised, chen2019temporal}. 
The setting of debiasing is similar to, but not the same as DA. 
The main difference is that we have no access to testing videos during training.
Besides, the debiasing setting does not assume the amount of testing data, 
while DA algorithms require a certain number of testing videos to determine the data distribution. 
\section{Formulation}
\label{sec:formulation}
Following~\cite{zhang2018mitigating}, the problem can be formulated as follows. 
For a video recognition dataset $D$, 
we can view each data sample as a tuple $(x, y, z)$ drawn from the joint distribution $ (X, Y, Z) $, 
where $x$ denotes the video, 
$y$ denotes the \emph{action} label, 
$z$ denotes one or multiple \emph{non-action} labels, 
such as \emph{scene}, \emph{object} or other attributes. 
We consider the supervised learning task, 
which builds a predictor $\hat{Y} = f(X)$ for $Y$ given $X$. 

Due to the dependence of $Y$ and $Z$ in the training set,
the predictors learned via standard supervised learning also 
yield predictions $\hat{Y}$ dependent on $Z$ given the action label $Y$.
Such behavior will lead to poor generalization capability, 
severely undermine the testing performance if $P(Z | Y)$ differs a lot between the train and test split.

Our goal is to learn non-discriminatory action recognition models \wrt $Z$,
which generalize well to testing videos 
with factors (\emph{scene}, \emph{object}, \eg) that rarely appear in the training set.
Non-dicrimination criterias have been of three types in fairness literature~\cite{barocas-hardt-narayanan}, 
\emph{independence} (\small $Y'\perp Z$\normalsize), 
\emph{seperation} (\small$Y'\perp Z|Y$\normalsize) 
and \emph{sufficiency} (\small$Y\perp Z|Y'$\normalsize). 
In the context of video recognition, we pursue \textsc{Equalized Odds}, similar to \emph{separation}, 
which is to minimizing the variance of \small$P(\hat{Y} = y | Y = y, Z = z)$ \normalsize for different $z$ given $y$. 
Besides improving the $z$-unbiasedness, 
we also need to maximize \small$P(\hat{Y} = y | Y = y)$ \normalsize to secure a good recognition model.

\begin{figure}[t]
	\centering
	\captionsetup{font=small}
	\includegraphics[width=.9\columnwidth]{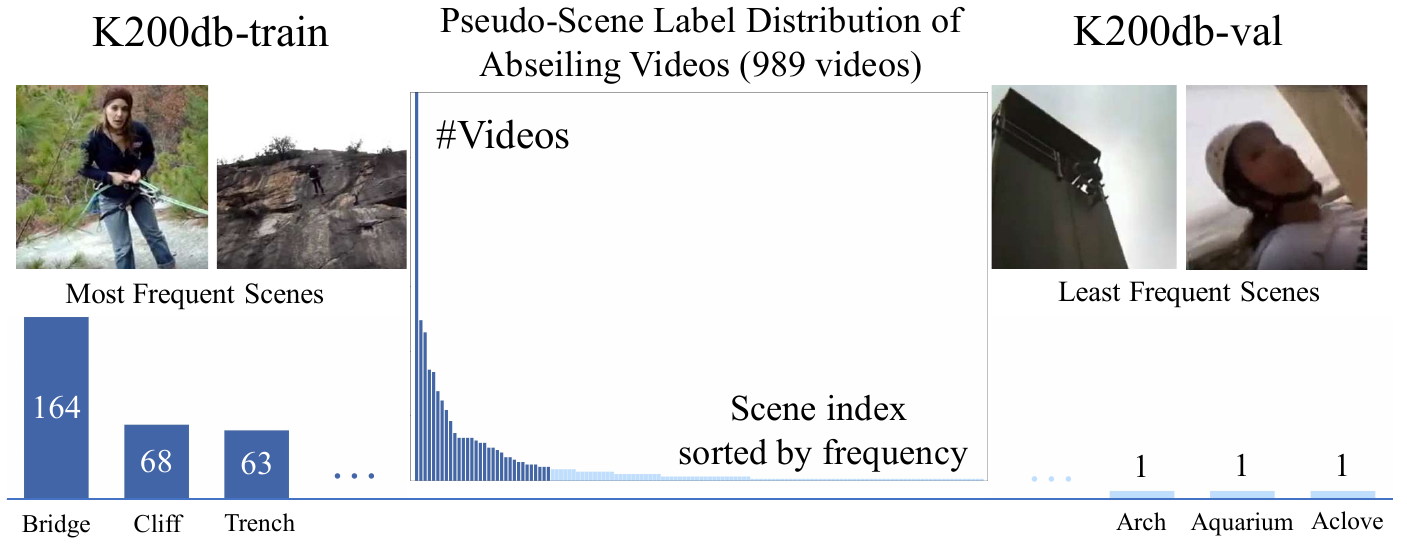} 
	\vspace{-2mm}
	\caption{\textbf{The long-tailed \emph{scene} distribution of abseiling videos.} 
		Most videos belong to several scene categories.
		In the distribution tail, there are many scene categories that rarely occur in training videos.
		We sample videos from the distribution head to form \texttt{K200db-train},
		from the tail to form \texttt{K200db-val}.}
	\label{fig-kinetics_split}
	\vspace{-5mm}
\end{figure}

\section{Evaluation Benchmark}
\subsection{Crafting Evaluation Datasets}
Most existing datasets assume the joint distribution $P(Y, Z)$ identical between train and validation splits.
To find out if an action recognition model is biased towards the \emph{non-action} labels, 
we design two evaluation protocols based on Kinetics-400~\cite{carreira2017quo}: 
re-distributing the existing train-val split and constructing a new validation set.

\begin{figure}[t]
	\centering
	\captionsetup{font=small}
	\includegraphics[width=.9\columnwidth]{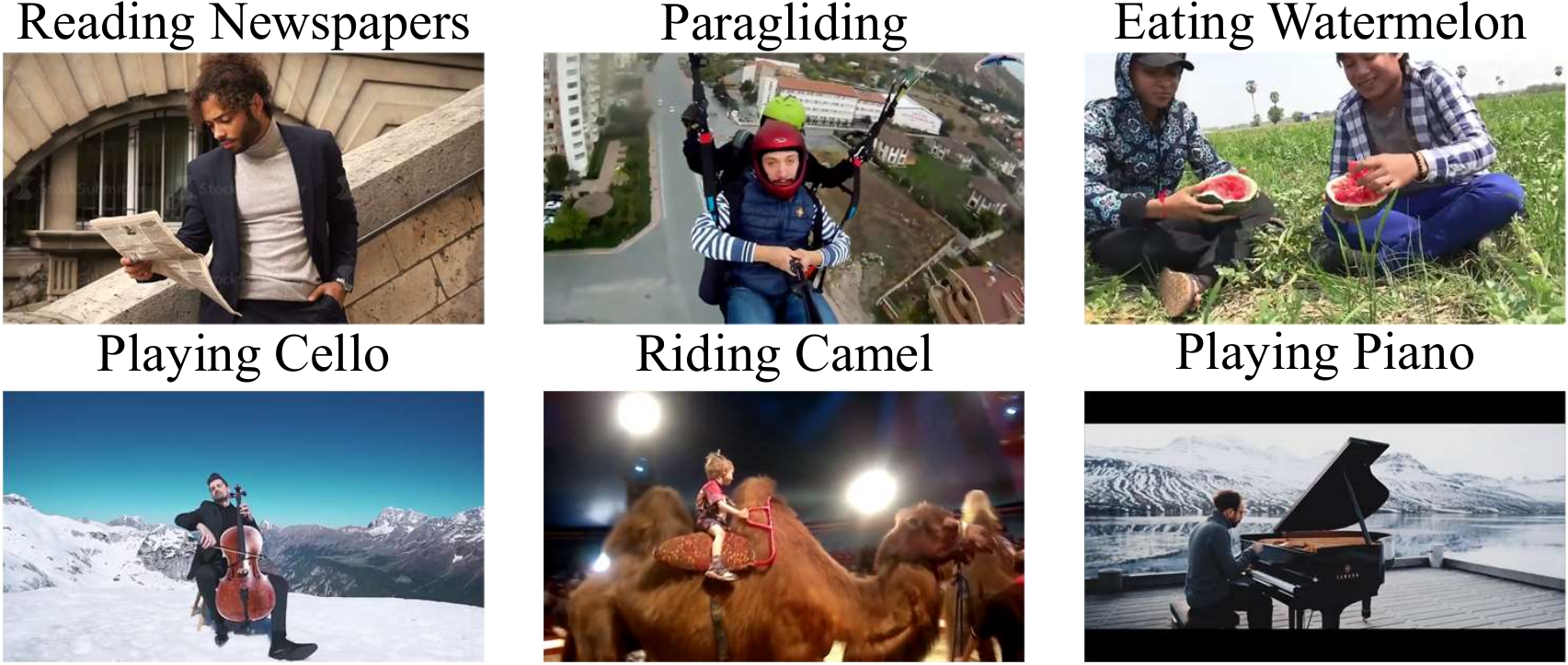} 
	\vspace{-2mm}
	\caption{\textbf{Samples of ARAS dataset.\protect\footnotemark}}
	\vspace{-3mm}
	\label{fig-ARAS}
\end{figure}

\begin{figure}[t]
	\begin{minipage}{.53\linewidth}
		\centering
		\captionsetup{font=small}
		\includegraphics[width=.9\textwidth]{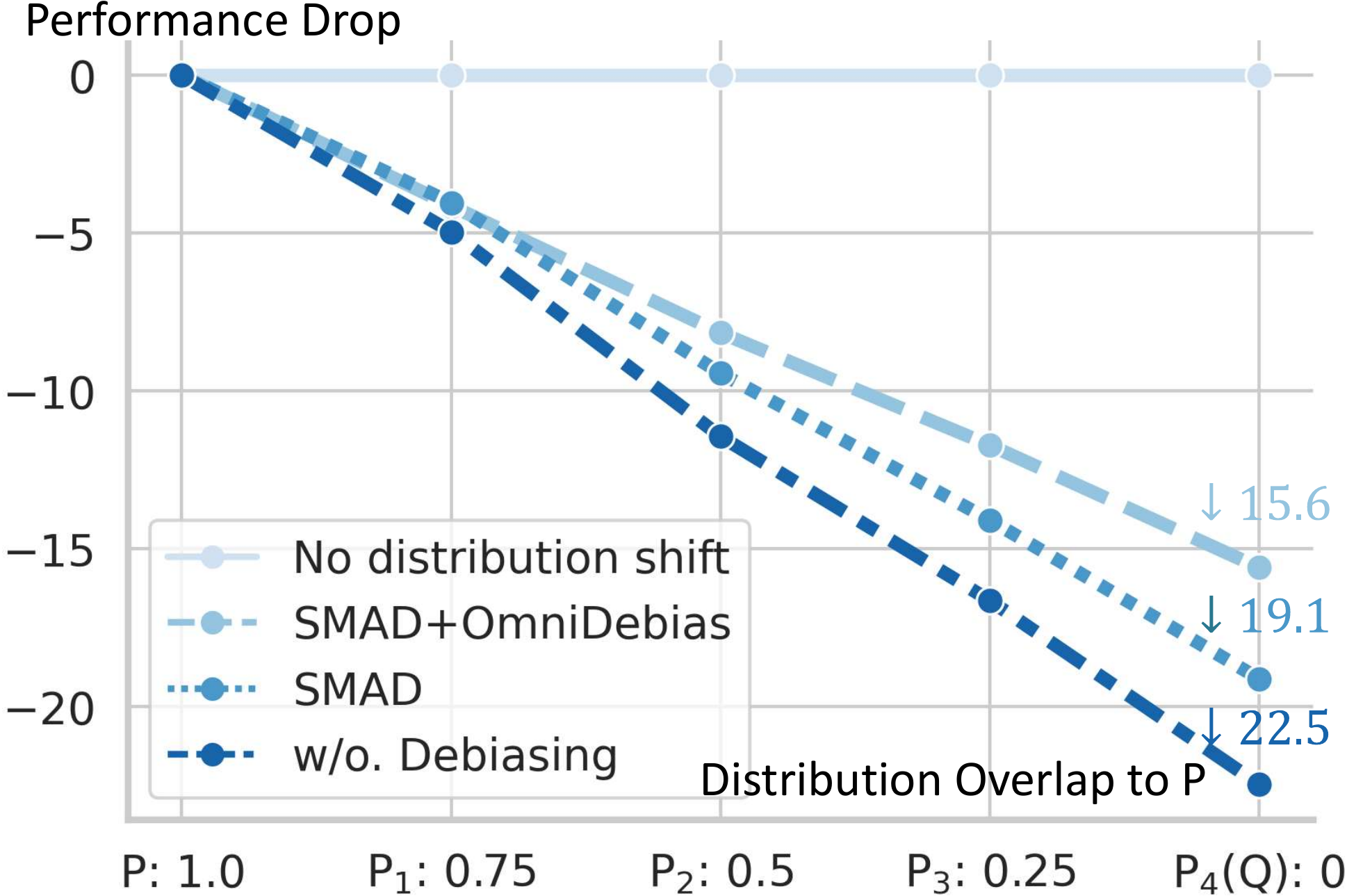}
		\vspace{-2mm}
		\captionof{figure}{ 
			\textbf{The Top1-Acc severely drops along with the scene distribution shift. }}
		\label{fig-overlap}
	\end{minipage}
	\hfill
	\begin{minipage}{.45\linewidth}
		\vspace{1.5mm}
		\footnotesize
		\centering
		\captionsetup{font=small}
		\resizebox{.9\linewidth}{!}{
		\tablestyle{12pt}{1.2}
		\begin{tabular}{c|cc}
			\shline
			TSN & Top1 Acc \\ 
			\shline
			\texttt{K200-val}		& 76.2	\\
			\texttt{K200-unbias}	& 53.7 \dec{22.5}	\\
			\texttt{ARAS-64}		& 55.8 \dec{20.4}	\\ 
			\shline
			SlowOnly & Top1 Acc \\ 
			\shline 
			\texttt{K200-val}		& 75.1	\\
			\texttt{K200-unbias}	& 51.9 \dec{23.2}	\\
			\texttt{ARAS-64}		& 51.0 \dec{24.1}	\\ 
			\shline 
		\end{tabular}}
		\vspace{-1mm}
		\captionof{table}{\textbf{Top1-Acc of TSN and SlowOnly on 3 test sets. }}
		\label{tab-evaluate}
	\end{minipage}
	\vspace{-4mm}
\end{figure}

\noindent\textbf{Re-distributing Train-Val Split.} 
We start with the original Kinetics-400 train split (with $\sim$240k videos). 
We apply a ResNet-50 trained on Places-365~\cite{zhou2017places} to obtain the pseudo \emph{scene} labels.
As shown in Figure~\ref{fig-kinetics_split}, 
the pseudo \emph{scene} labels have a long-tailed distribution.
We take the tail as the validation set and sample a subset from the head to be the training set.
To maintain the inter-class sample balance, 
we select 200 classes with the most training samples and construct a subset that contains 80k videos for training and 10k for validation,
(denoted as \texttt{K200db-train} and \texttt{K200db-val}, \texttt{db} for debiasing).
We examine the \emph{action-scene} correlation of the two splits by calculating the 
normalized mutual information (NMI) of \emph{action} and \emph{scene}: 
for \texttt{K200db-train}, the NMI is $0.466$ ($0.397$ if sampled randomly);
for \texttt{K200db-val}, the NMI is $0.374$ ($0.488$ if sampled randomly).
Based on the splitting method,
we can also tune the overlap of common \emph{scene} labels in \texttt{K200db-train} and \texttt{K200db-val} 
for varying distribution shift.

\noindent\textbf{Constructing New Validation Set.}
Beyond being restricted to the original dataset, we can further construct a new dataset for evaluation.
This resembles the real-world scenario: the trained model is fixed while the environment changes at deployment.
We begin with \emph{action} labels in Kinetics and consider some \emph{rare scenes}.
The combinations of \emph{actions} and \emph{rare scenes} are used as queries to obtain web videos from YouTube. 
We manually examine the web videos and obtain around ten videos for each class in 104 Kinetics classes, 
denoted as Action with RAre Scenes (\texttt{ARAS-104}).
For \texttt{K200db}, there are 64 overlapped classes (\texttt{ARAS-64}).
Figure~\ref{fig-ARAS} shows several examples.
We use \texttt{ARAS} to simulate the out-of-distribution testing 
for \emph{scene}-debiasing evaluation.

\footnotetext{ARAS video samples in: \url{https://youtu.be/j1LA3y-UuEA}. }

\begin{figure*}[t]
	\centering
	\captionsetup{font=small}
	\includegraphics[width=.95\textwidth]{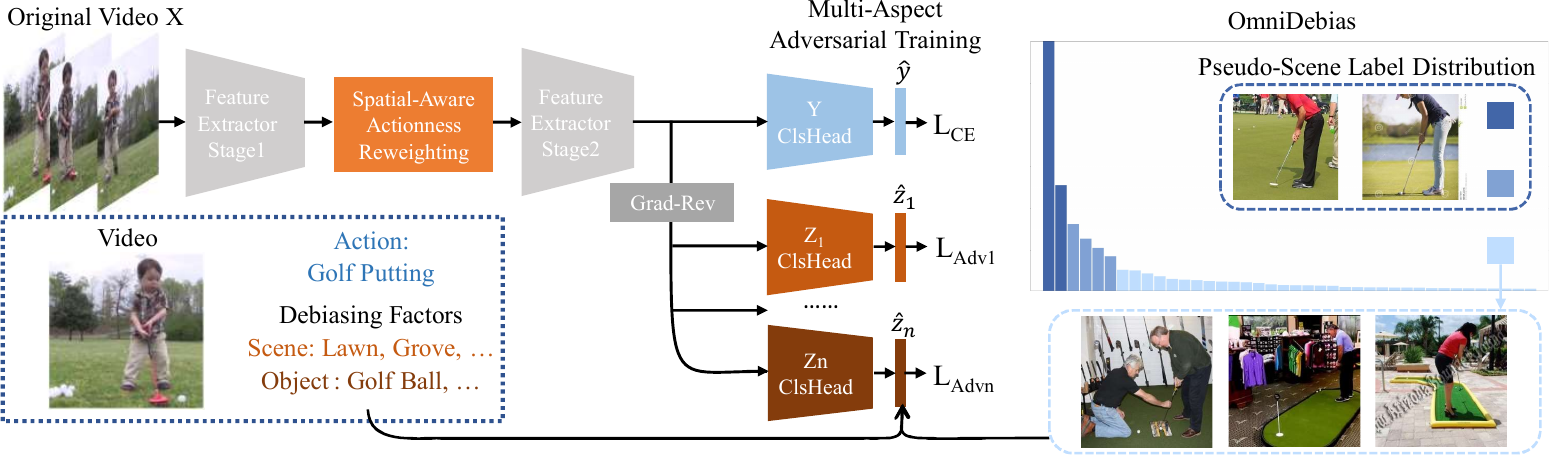}
	\vspace{-2mm}
	\caption{\textbf{SMAD \& OmniDebias.} 
		\textbf{Left: SMAD framework}.
		Multiple adversarial heads are used in SMAD for Multi-Aspect debiasing.
		The SAAR module (Figure~\ref{fig-implicit}) is inserted in the backbone to improve the spatial modeling capability.
		\textbf{Right: OmniDebias}.
		OmniDebias only uses the \texttt{unbias} part of web media for joint training,
		achieving better performance and efficiency.
	}
	\label{fig-ei_debiasing}
	\vspace{-5mm}
\end{figure*}

\subsection{Evaluation of Existing Methods }
We first evaluate existing methods on the new benchmarks, 
including a 2D-CNN method (TSN-3seg-R50)~\cite{wang2016temporal} and 
a 3D-CNN one (SlowOnly-8x8-R18)~\cite{feichtenhofer2019slowfast}.
From Table~\ref{tab-evaluate}, we observe that 
the Top-1 accuracies on both \texttt{K200db-val} and \texttt{ARAS-64} are 
significantly lower than the original validation split \texttt{K200-val}.
This reflects models learned with vanilla training cannot handle the large discrepancy of the \emph{action-scene} joint distribution between train/val splits.
\texttt{K200db-[train/val]} is an extreme case that has disjoint scene labels.
We can also vary the overlap of scene labels between \texttt{K200db-val} and \texttt{K200db-train}.
Figure~\ref{fig-overlap} demonstrates that the drop of accuracy is positively correlated to the distribution shift.
That performance drop can be largely mitigated by \textbf{SMAD} and \textbf{OmniDebias}, 
which will be detailed in the following section.

\section{Method}
We devise Spatial-aware Multi-Aspect Debiasing (\textbf{SMAD})
which seeks to learn a representation invariant to multiple aspects of videos, 
\eg, \emph{scene}, \emph{object}, and other attributes, with adversarial training.
Besides, we propose OmniDebias to harness the richness and diversity of web data efficiently,
to improve the expressive power of the learned representation.
We integrate the two complementary aspects into a unified framework, 
as illustrated in Figure~\ref{fig-ei_debiasing}.

\subsection{Spatial-aware Multi-Aspect Debiasing}
SMAD incorporates both \textbf{explicit} debiasing using Multiple Aspects as Adversarial Training objectives (MAAT) 
and \textbf{implicit} debiasing with Spatial-Aware Actionness Reweighting (SAAR).

 \textbf{Multi-Aspect Adversarial Training.}
We denote each input as a tuple $(x, y, z_1, \cdots, $ $z_M) \in \cX \times \cY \times \cZ_1 \times \cdots \times \cZ_M $, 
where we pre-define $ M $ aspects in addition to the set of \emph{action} labels $ \cY $.
We use a ConvNet $f_\Theta$ parameterized by $ \Theta $ for feature extraction.
On top of $f_\Theta$ are $(M+1)$ classification heads:
one head $h_\cY$ (parameterized by $\theta_{\cY}$) to predict the \emph{action} $y$ and 
$M$ adversarial heads $h_{\cZ_i}$ (parameterized by $\theta_{\cZ_i}$) to recognize tags belong to the aspect $\cZ_i$.
We use the standard cross-entropy loss $L_{\mathrm{ce}, \cY}$ to train $h_\cY$, 
use adversarial losses $L_{\mathrm{adv}, \cZ_i}$ for the rest \emph{non-action} heads $h_{\cZ_i}$.

The optimization can be divided into two parts: classification heads and the backbone.
For classification heads, 
the objective is to minimize $L_{\mathrm{ce}, \cY}$ and $L_{\mathrm{adv}, \cZ_i}$:
\begin{equation}
\small
\theta_{\cY}, \theta_{\cZ_1}, \cdots \theta_{\cZ_M} = \argmin_{\theta_{\cY}, \theta_{\cZ_1}... \theta_{\cZ_M}} ( L_{\mathrm{ce}, \cY} + \sum_{i=1}^{i=M} \lambda_i L_{\mathrm{adv}, \cZ_i}).
\end{equation}
$\lambda_i$ is the weight of the adversarial loss.
For the backbone, since we aim for feature that is both discriminative for $\cY$ and invariant for $Z_i$, 
the objective is to minimize $L_{\mathrm{ce}, \cY}$ and \emph{maximize} $L_{\mathrm{adv}, \cZ_i}$: 
\begin{equation}
\small
\Theta = \argmin_{\Theta} (L_{\mathrm{ce}, \cY} - \sum_{i=1}^{i=M} \lambda_i L_{\mathrm{adv}, \cZ_i}).
\end{equation}

By inserting a gradient reversal layer~\cite{ganin2015unsupervised} before $h_{\cZ_1}, \cdots, h_{\cZ_M} $, 
we can simultaneously optimize the backbone $f_\Theta$ along with all heads
efficiently using the standard stochastic gradient descent.

\textbf{Choice of Adversarial Losses.}
The type of the adversarial loss depends on the label format of $\cZ_i$.  
For soft-label $\cZ_i$, we can use soft cross-entropy loss (\textbf{SoftCE}, Eq.~\ref{eq:softce}) 
or KL-divergence loss (\textbf{KLDiv}, Eq.~\ref{eq:kldiv}).
For multi-label $\cZ_i$, we use binary cross-entropy loss (\textbf{BCE}). 
\begin{equation}
\small
L_{\mathrm{adv}, \cZ_i} = - \sum_{k=1}^{k=\vert \cZ_i \vert } z_{i_k} \log [h_{\cZ_i}(f(x;\Theta))]_k.
\label{eq:softce}
\end{equation}
\begin{equation}
\small
L_{\mathrm{adv}, \cZ_i} = \sum_{k=1}^{k=\vert \cZ_i \vert } z_{i_k} \log \frac{z_{i_k}}{[h_{\cZ_i}(f(x;\Theta))]_k}.
\label{eq:kldiv}
\end{equation}
\vspace{1mm}

\textbf{Source of Non-Action Labels.}
The labels for $\cZ_1, \cdots, \cZ_M$ are needed in adversarial training.
However, these annotations are usually unavailable for most action recognition datasets.
To handle this, we use off-the-shelf ConvNets trained on the specific domain to obtain the pseudo labels.
For example, we use ResNet trained on ImageNet~\cite{deng2009imagenet} and Places365~\cite{zhou2017places} to 
assign the pseudo labels for \emph{object} and \emph{scene}, respectively.

\begin{figure}[t]
	\centering
	\captionsetup{font=small}
	\includegraphics[width=.65\columnwidth]{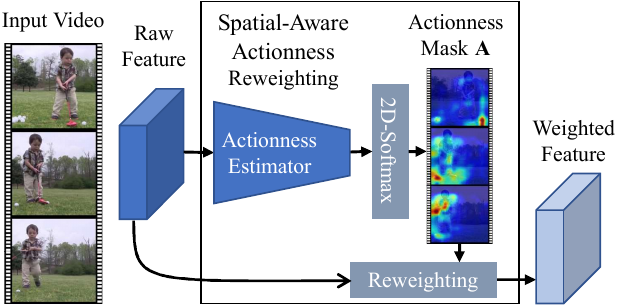}
	\vspace{-1mm}
	\caption{\textbf{SAAR module.} 
		The Spatial-Aware Actionness Reweighting module learns an actionness mask $ \mA $ to reweight features at different locations. 
		The learned mask has low values on irrelated scenes or objects to suppress these features. }
	\label{fig-implicit}
	\vspace{-4mm}
\end{figure}

\textbf{Spatial-Aware Actionness Reweighting. }
The idea of adversarial training is simple but turns out to be fragile, 
especially when considering multiple aspects.
It would be hard for the vanilla algorithm to converge 
if the adversarial loss weight $\lambda_i$ is set as a relatively large number.
One possible conjecture is that the underlying network pools feature \emph{uniformly} across all positions.
Since we can only mitigate the model bias instead of eliminating the dataset bias, 
the inherent bias from the data would contradict the adversarial objective unless the model selectively attends to the action-related region.
To this end, inspired by the idea of actionness estimation~\cite{wang2016actionness,zhao2017temporal}, 
we propose a Spatial-Aware Actionness Reweighting module (SAAR), 
illustrated in Figure~\ref{fig-implicit}.

For a feature map $\mF \in \mR^{C \times T \times H \times W}$, 
we first estimate an actionness mask $ \mA \in \mR^{T \times H \times W}$, 
where the scalar for each location represents how much the feature is related to the human action.
In experiments, we use a 2D ResNet-Layer with a small bottleneck width for actionness feature extraction, 
and use another 2D 3\x 3 convolution as the actionness head, 
which outputs a 1-channel actionness score map $\mA$.
On top of the score map, we apply 2D-softmax across the spatial dimensions for normalization:
\small$ \mA'(t, h, w) = \frac{e^{\mA(t, h, w)}}{\sum_{h',w'} e^{\mA(t, h', w')}}$ \normalsize.
The final modulated feature map is the element-wise multiplication between $\mF$ and $\mA'$: 
\begin{equation}
\small
\mF'(c,t,h,w) = (H\times W)\cdot \mF(c,t,h,w) \odot \mA'(t,h,w)
\end{equation} 
where the coefficient $ H\times W $ is used to preserve the magnitude of feature maps after re-weighting.
We insert SAAR before the last ResNet-Layer in the backbone.
Operating on a small feature map (14\x 14), the SAAR module adds up to 2\% additional computation. 

Experiments show that spatial-aware actionness reweighting can 
not only benefit convergence of training but also lead to better performance.
It is worth noting that the benefit of SAAR is much larger when combined with MAAT than used alone, 
indicating that the adversarial training objective incurs weak supervision implicitly.

\subsection{Exploiting Web Media with OmniDebias}

Instead of restricting to labeled datasets,
we also propose to leverage webly-supervised datasets for bias mitigation via co-training, 
considering their richness and diversity.

We use \texttt{GoogleImg} (\texttt{GG}) and \texttt{InsVideo} (\texttt{IG}) from the 
OmniSource dataset~\cite{duan2020omni} as the web data source. 
Following the same pipeline as the original work,
to construct the auxiliary dataset for joint training,
we train a teacher network to filter web data and keep high-confidence examples.
Joint training with the built auxiliary dataset 
can lead to much larger improvements on our evaluation benchmarks (\texttt{K200db-val}, \texttt{ARAS}), 
compared to the improvement on standard validation sets (\texttt{K200/400-val}),
mostly because web media contain novel $z\in\cZ$ that does not exist in the training set.

\begin{figure}[t]
	\centering
	\captionsetup{font=small}
	\includegraphics[width=.9\columnwidth]{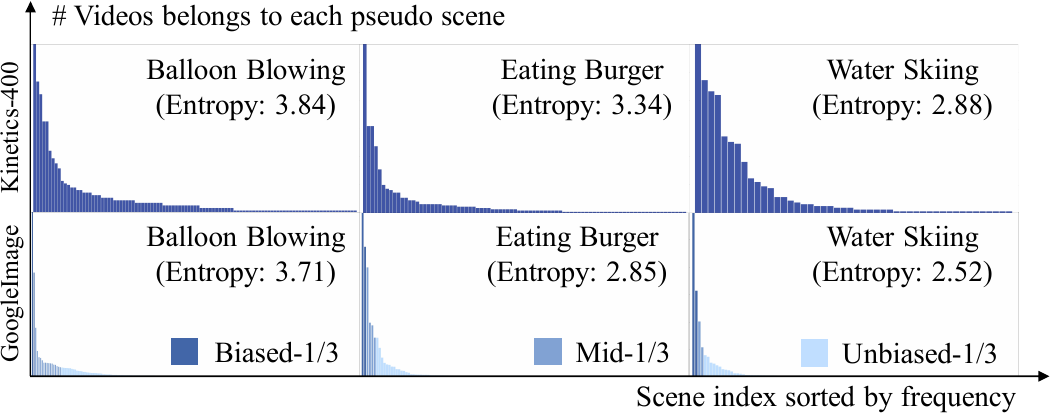}
	\vspace{-3mm}
	\caption{\textbf{Pseudo-scene distributions. } 
	Visualization of pseudo scene distributions of 3 action categories in Kinetics-400 and GoogleImg. 
	GoogleImg has more imbalanced distributions.}
	\label{fig-dist_comp}
	\vspace{-2mm}
\end{figure}

However, there is a drawback to the naïve approach.
For web data, the distribution over $z$ can be even more imbalanced than the distribution for Kinetics videos.
For example, the average entropy of pseudo \emph{scene} distributions of 400 actions is 3.02 for Kinetics, 
and 2.79 for \texttt{GoogleImg} (larger $\rightarrow$ more diversified).
Figure~\ref{fig-dist_comp} demonstrates pseudo \emph{scene} distributions of 3 action classes.
Co-training with such an unbalanced dataset is sub-optimal.

Thus we propose OmniDebias to utilize web media more efficiently.
In OmniDebias, we use a simple data selection strategy to select a subset
of the entire web dataset for joint training.
Specifically, based on the same approach introduced in \textbf{Benchmark}, 
we sort the samples in a same action class by the descending order of $z$-frequency
\footnote{For $z$ = \texttt{scene}, 
	if 20 out of 100 samples have the scene label `cliff', 
	the $z$-frequency of each of the 20 samples is 0.2 (20 / 100). }.  
Based on the $z$-frequency, 
we divide the auxiliary dataset into 3 equal-sized parts,
\ie \texttt{[web]-bias}, \texttt{[web]-mid} and \texttt{[web]-unbias} 
(\texttt{[web]} can be \texttt{GG}, \texttt{IG}, \etc), 
and use \texttt{[web]-unbias} only for joint training.
OmniDebias consistently outperforms not only using other parts but also the union, 
indicating its efficacy and efficiency.

\section{Experiments}

\begin{table*}[t]
	\centering
	\captionsetup{font=small}
	\footnotesize
	\resizebox{\linewidth}{!}{
		\tablestyle{4pt}{1.3}
		\begin{tabular}{cc|ccc|ccc|ccc}
			\shline
			\multicolumn{2}{c|}{Component}	& \multicolumn{3}{c|}{SMAD}	& \multicolumn{3}{c|}{OmniDebias}	& \multicolumn{3}{c}{Combination}	\\ 
			\shline
			Model & Train / Test	& ARAS-64	& K200db-val	& K200-val	& ARAS-64	& K200db-val & K200-val	& ARAS-64	& K200db-val	&K200-val	\\ 
			\shline
			2D    & K200db-train	& 60.3 \inc{4.5} & 55.7 \inc{2.0} & 75.5 \dec{0.7} & 68.3 \inc{12.5} & 60.0 \inc{6.3} & 77.4 \inc{1.2} & 70.9 \inc{15.1} & 62.0 \inc{8.3}  & 78.1 \inc{1.9} \\
			3D    & K200db-train	& 58.4 \inc{7.4} & 55.0 \inc{3.1} & 74.6 \dec{0.5} & 69.2 \inc{18.2} & 60.7 \inc{8.8} & 78.7 \inc{3.6} & 71.6 \inc{20.6} & 62.7 \inc{10.8} & 78.4 \inc{3.3} \\
			\shline
			Model & Train / Test & ARAS-104    & -           & K400-val    & ARAS-104     & -           & K400-val    & ARAS-104     & -            & K400-val    \\ 
			\shline
			2D    & K400-train                & 56.2 \inc{2.1} & -           & 70.0 \dec{0.6} & 60.2 \inc{6.1}  & -           & 71.3 \inc{0.7} & 61.8 \inc{7.7}  & -            & 71.4 \inc{0.8} \\
			3D    & K400-train                & 55.0 \inc{3.5} & -           & 68.2 \dec{0.1} & 60.2 \inc{8.7}  & -           & 71.0 \inc{2.7} & 61.2 \inc{9.7}  & -            & 70.8 \inc{2.5} \\
			\shline
		\end{tabular}}
	\vspace{2mm}
	\caption{
		\textbf{The individual and joint effects of SMAD and OmniDebias.} 
		We report the Top-1 accuracies on three test sets: \texttt{ARAS}, 
		\texttt{K200db-val} and \texttt{K200/400-val}. 
		$\uparrow$ and $\downarrow$ denote the improvement or decline to the baseline w/o. debiasing.
	}
	\label{tab-main}
	\vspace{-2mm}
\end{table*}

\begin{table}[t]
	\begin{minipage}{.49\linewidth}
		\centering 
		\captionsetup{font=small}
		\footnotesize
		\resizebox{\linewidth}{!}{
			\tablestyle{8pt}{1.3}
			\begin{tabular}{cccc}
				\shline
				Method          								& Access 	& K200db-val& ARAS-64\\ 
				\shline
				Baseline        								& \xmark   	& 51.9  	&  51.0	\\ 
				\shline
				AdaBN \cite{li2016revisiting} 			& \xmark   	& 52.0 		&  52.3 \\
				FrameShuffle \cite{Carter2020AnAF}  	& \xmark   	& 52.4 		&  51.3	\\
				SDN \cite{choi2019can} 					& \xmark 	& 54.0 		&  55.3	\\
				\shline
				DANN \cite{ganin2015unsupervised}  		& \cmark   	& 52.4  	&  53.3	\\
				MMD \cite{gretton2012kernel}   			& \cmark   	& 52.7  	&  53.1	\\
				SAVA \cite{choi2020shuffle} 			& \cmark   	& 53.4 		&  54.0	\\
				\shline
				SMAD  & \xmark   	& 55.0  	&  58.4	\\ \shline
			\end{tabular}}
		\vspace{2mm}
		\captionof{table}{
			\textbf{SMAD v.s. other debiasing and domain-adaptation algorithms. } 
			\textbf{Access} indicates if the algorithm needs to access validation data in training.
		}
		\label{tab-da}
	\end{minipage}
	\hfill
	\begin{minipage}{.49\linewidth}
		\centering
		\captionsetup{font=small}
		\footnotesize
		\resizebox{\linewidth}{!}{
			\tablestyle{6pt}{1.3}
		\begin{tabular}{cccc}
			\shline
			Metric & \begin{tabular}[c]{@{}c@{}}Independence\\  $I(Y', S)$ \end{tabular} & \begin{tabular}[c]{@{}c@{}}Separation\\ $I(Y', S|Y)$\end{tabular} & \begin{tabular}[c]{@{}c@{}}Sufficiency\\ $I(Y, S|Y')$\end{tabular} \\ 
			\shline
			Baseline    & 0.498 \rinc{0.011}	& 0.376 & 0.356  \\
			SMAD        & 0.490 \rinc{0.003}	& 0.373 & 0.366  \\
			OmniDebias  & 0.496 \rinc{0.009}	& 0.334 & 0.321  \\
			Combination & 0.491 \rinc{0.004}	& 0.321 & 0.327  \\ 
			\shline
		\end{tabular}}
		\vspace{1mm}
		\caption{\textbf{Fairness metrics on K200-val. } 
			We train the SlowOnly-R18-8x8 on \texttt{K200db-train}, 
			$I$ denotes normalized mutual information
			(lower $\rightarrow$ more non-discriminative). 
			\red{Red} marks denote the \emph{scene}-bias amplified to the oracle independence $I(Y, S) = 0.487$. }
		\label{tab-fairness}
	\end{minipage}
\end{table}

\subsection{Experiment Setting}
\noindent\textbf{Acquisition of non-action labels.}
For debiasing, \emph{non-action} labels can be either pseudo labels inferred by a pretrained model 
or ground-truth labels from a multi-label dataset.
To acquire pseudo labels for debiasing, 
we use ResNet50~\cite{he2016deep} pretrained on ImageNet and Places365 as the pseudo label extractor
for \emph{scene} and \emph{object}.
We also tried ResNet18 and DenseNet161~\cite{huang2017densely} as the extractor for \emph{scene} labels
but observe a subtle difference ($\le 0.3\%$). 
For ground-truth labels, we use the HVU dataset~\cite{diba2019holistic}, 
which annotates Kinetics videos with three additional tag categories: \emph{context}, \emph{attribute}, \emph{event}.

\noindent\textbf{Training and Evaluation.}
We use TSN-3seg-R50 with ImageNet pretraining as the 2D-CNN baseline, 
SlowOnly-8x8-R18 as the 3D-CNN one.
In the choice of adversarial losses, 
we use a weighted combination of SoftCE and KLDiv for soft-label (better than each individual), 
use BCE for multi-label. 
The loss weight is 0.5 for SoftCE and KLDiv losses, 5 for BCE loss.
For testing, we uniformly sample 25 frames for TSN or ten clips for SlowOnly with center crop and average the final predictions.
In experiments, we report the Top-1 accuracy.
Since ARAS is a small evaluation set, we first examine the statistical significance:
we train the TSN on \texttt{K200db-train} for 10 times with different random seeds and test it on \texttt{ARAS-64}.
We find the standard deviation of accuracy is around $ 0.3\% $,
which means a difference larger than 0.8\% is statistically significant.

\subsection{Main Results}
\noindent \textbf{Re-distributed Train-Val Splits.}
When the training and validation subsets have different scene distributions,
our method consistently bridges the performance gap between validation sets with \emph{scene} distribution shift and
the validation set without \emph{scene} distribution shift, as shown in~Figure~\ref{fig-overlap}.
The narrower accuracy gaps reflect the improvement made under the fairness metrics \textsc{Equalized ODDs}. 
When the testing and training scene distributions are completely different, \ie disjoint label sets,
the effect is most significant: our methods reduce the accuracy drop by nearly $\frac{1}{3}$: from 22.5\% to 15.6\%.

Besides \textsc{Equalized ODDs}, we also evaluate three commonly used fairness metrics,
namely \emph{independence}, \emph{separation}, and \emph{sufficiency} in Table~\ref{tab-fairness}.
SMAD largely mitigates the bias amplified by the algorithm:
without SMAD, $I(Y', S)$ increases 0.011 (the \emph{scene}-bias amplified by algorithm) 
compared to $I(Y, S)=0.487$,
while SMAD reduces it to 0.003.
Since $Y$, $S$ are not statistically independent, 
the sufficiency and independence cannot both hold.
Thus we observe that $I(Y, S|Y')$ increases when we apply SMAD for debiasing.
For OmniDebias, since additional web media are used for joint training,
all three fairness metrics are improved (lower $\rightarrow$ better). 

We further study the individual and combined effects of SMAD and OmniDebias.
Extensive experiments are conducted with both 2D and 3D baselines: 
models are trained on \texttt{K200db-train} or \texttt{K400-train} and tested on 3 test sets: 
\texttt{ARAS-64/104}, \texttt{K200db-val} ($z$-unbiased) and \texttt{K200/400-val} (normal).
The results are demonstrated in Table~\ref{tab-main}. 
SMAD can improve the performance on $z$-unbiased test sets by a large margin 
at the cost of a little accuracy drop on the normal test set.
The improvement of OmniDebias is across all 3 test sets 
since additional web media are used for joint training.
while for \texttt{K200db-val} and \texttt{ARAS} the gain is much more noticeable.
Combining SMAD and OmniDebias yields the highest accuracy on all $z$-unbiased test sets, 
indicates that the two techniques are orthogonal to each other.

\noindent \textbf{A new Debiasing Benchmark. }
In Table~\ref{tab-da}, we evaluate multiple debiasing and domain-adaptation algorithms 
on our new \textbf{facet-based re-distribution} and \textbf{out-of distribution} benchmarks. 
The models (backbone: SlowOnly-R18) are trained on \texttt{K200db-train}, tested on \texttt{K200db-val} and \texttt{ARAS-64}. 
SMAD is a better solution for the debiasing problem compared to the alternatives,   
considering its superior performance and simple deployment. 

\noindent \textbf{Transferring Abilities.}
The debiased representation is also more useful when transferred to other tasks.
We study two cases: few-shot learning and video classification.
SlowOnly-8x8-R18 trained on \texttt{K200db-train} is used as the feature extractor.
For each video, we uniformly sample 10 clips, 
extract a 512-d feature for each clip, 
and concatenate them into a 5120-d video-level feature.

We evaluate the few-shot learning performance on FineGYM-99~\cite{shao2020finegym}, 
a fine-grained gymnastic action recognition dataset with less scene bias.
We construct 10,000 5-way episodes (1-shot or 5-shot).
In each episode, the cosine similarities between the query sample and support samples are used for classification.
Table~\ref{tab-downstream} shows that both SMAD and OmniDebias contribute to the few-shot performance on FineGYM-99.

\begin{table}[t]
	\centering
	\captionsetup{font=small}
	\scriptsize
	\resizebox{\columnwidth}{!}{
		\tablestyle{10pt}{1.2}
		\begin{tabular}{c|cc|ccc}
			\shline
			Setting  & GYM-1shot & GYM-5shot & HMDB51 & UCF101 & Diving48 \\ 
			\shline
			w/o. Debiasing   	& 42.2   & 52.9   & 49.9   & 84.3   & 17.3   \\ 
			\shline
			+ SMAD     			& 45.5   & 56.4   & 50.9   & 84.9   & 18.9   \\ 
			+ IG-all   			& 46.6   & 58.4   & 54.3   & 88.4   & 19.8   \\
			+ IG-unbias			& 47.1   & 59.5   & 55.9   & 88.8   & 20.9   \\ 
			\shline
			+ SMAD, IG-unbias 	& 51.7   & 62.1   & 57.2   & 89.9   & 22.3   \\ 
			\shline
	\end{tabular}}
	\vspace{2mm}
	\caption{\textbf{Few-shot Learning \& Feature Classification. }
		The learned representation achieves good performance on downstream tasks. 
		We report the 3-split average for HMDB51 and UCF101.}
	\label{tab-downstream}
\end{table}

We evalute the performance of video classification on 
UCF101~\cite{soomro2012ucf101}, 
HMDB51~\cite{kuehne2011hmdb} 
and Diving48~\cite{li2018resound} with two settings: 
feature classification and finetuning.
For feature classification,
we train a linear SVM based on the 5120-d descriptors.
Table~\ref{tab-downstream} shows that both SMAD and OmniDebias improve the feature classification performance.
Two baselines are used in the finetuning setting.
We first use ResNet3D-18~\cite{hara2018can} trained on MiniKinetics~\cite{xie2017rethinking} 
with input size 112 as the baseline, 
for a straightforward comparison with SDN~\cite{choi2019can} (Table~\ref{tab-finetune} upper).
With pseudo labels for recognition only (much cheaper than pseudo labels for human detection), 
SMAD can outperform SDN on three downstream tasks.
By introducing web data with OmniDebias, 
the model can obtain much better performance.
We further test the finetuning performance on SlowOnly-8x8-R18 trained with K200db-train, 
which is the setting used across this paper (Table~\ref{tab-finetune} lower).
The improvement of SMAD and OmniDebias is also steady and distinct upon this much stronger baseline.

\begin{table}[t]
	\centering
	\captionsetup{font=small}
	\resizebox{\columnwidth}{!}{
		\tablestyle{10pt}{1.2}
		\begin{tabular}{c|c|cccc}
			\shline
			Method          & Pretrain 	& HMDB51	& UCF101	& Diving48*	 & Diving48 \\ 
			\shline
			ResNet3D-16x1 	& MiniKinetics 	& 53.6   	& 83.5   	& 18.0 & - 	\\ 
			+ SDN~\cite{choi2019can}& MiniKinetics	& 56.7  & 84.5   	& 20.5 & - 	\\ 
			+ SMAD			& MiniKinetics  & 57.2		& 84.7		& 20.9 & - 	\\ 
			+ SMAD, IG-unbias	& MiniKinetics  & 61.2		& 88.2		& 22.6 & -  \\ 
			\shline
			SlowOnly-8x8	& K200db-train     & 62.6   	& 89.6   	& 25.5 & 53.9  \\ 
			+ SMAD          & K200db-train     & 64.0 		& 90.4		& 26.7 & 55.7  	\\
			+ SMAD, IG-unbias  	& K200db-train & 67.3		& 93.3		& 28.2 & 59.7  	\\ 
			\shline 
	\end{tabular}}
	\vspace{2mm}
	\caption{
		\textbf{Finetuning performance.} 
		Our work improves the finetuning performance on 3 datasets significantly 
		under different settings.
		We report the 3-split average for HMDB51 and UCF101. 
		* denotes using the old version of Diving48 annotations. }
	\label{tab-finetune}
	\vspace{-4mm}
\end{table}

\subsection{Spatial-aware Multi-Aspect Debiasing }

\begin{table}[t]
	\centering
	\captionsetup{font=small}
	\resizebox{.7\linewidth}{!}{
		\tablestyle{8pt}{1.3}
	\begin{tabular}{c|cccc}
		\shline
		Test Set  & Baseline & MAAT  & SAAR  & MAAT + SAAR \\ 
		\shline
		ARAS-64   & 51.0     & 55.8  & 51.6  & 58.4     \\ 
		K200db-val & 51.9    & 54.3  & 52.7  & 55.0     \\ 
		\shline
	\end{tabular}}
	\vspace{2mm}
	\caption{\textbf{Performance of MAAT \& SAAR.}
		The baseline is SlowOnly-8x8-R18 trained on \texttt{K200db-train}. }
	\label{tab-eidebias}
\end{table}

\noindent\textbf{Ablation of SMAD.}
We first evaluate the efficacy of two components in SMAD, 
namely Multi-Aspect Adversarial Training (MAAT) and Spatial-Aware Actionness Reweighting (SAAR).
Table~\ref{tab-eidebias} demonstrates that MAAT itself can largely improve the performance on test videos with novel scenes (\texttt{ARAS-64}, \texttt{K200db-val}).
Upon this decent baseline, SAAR further boost the performance by 0.7\% on \texttt{K200db-val} and 2.6\% on \texttt{ARAS-64}.
The improvement is non-trivial since SAAR only introduces 2\% additional FLOPs and requires no additional explicit supervision.
It is also worth noting that without the guidance from MAAT, the gain of SAAR is much reduced. 
The combination of MAAT and SAAR achieves large improvement on videos with novel scenes. 

\begin{table}[t]
	\centering
	\captionsetup{font=small}
	\resizebox{.65\linewidth}{!}{
	\tablestyle{10pt}{1.2}
	\begin{tabular}{c|c}
		\shline
		Debias Factor	& K200db-val  	\\		
		\shline
		None			& 51.7				\\ 	
		\shline
		\emph{scene}			& 53.2 \inc{1.5} 	\\
		\emph{object}			& 52.9 \inc{1.2} 	\\	
		\shline
		\emph{scene, object}	& 53.5 \inc{1.8} 	\\	
		\emph{scene, object, event, attribute, context} & 53.8 \inc{2.1} \\
		\shline
		\end{tabular}}
	\vspace{2mm}
	\caption{\textbf{Multi-factor v.s. single-factor debiasing.}}
	\label{tab-universal}
	\vspace{-2mm}
\end{table}

\noindent \textbf{Advantages of Multi-Aspect Debiasing.}
Multi-aspect debiasing is more generic than the \emph{scene}-debiasing algorithm~\cite{choi2019can}.
To prove that, we design a complex dataset split 
(\texttt{K200-both-split}, split by both \emph{scene} and \emph{object}) 
to mimic the real-world debiasing scenario.
Specifically, we first create \texttt{K200-scene-split} and \texttt{K200-obj-split} 
using the introduced re-distributing method,
with the factor \emph{scene} and \emph{object} respectively.
Then we sample the validation videos from the union of two validation sets
and sample the training videos from the remaining videos to form \texttt{K200-both-split}. 
On that split, we evaluate different debiasing factors. 
For multi-aspect debiasing with $N$ factors, the weight of each adversarial loss is divided by $N$.  
Table~\ref{tab-universal} shows that multi-aspect debiasing consistently outperforms the single-aspect one 
under this setting: 
using both \emph{scene} and \emph{object} as debiasing factors outperforms each individual. 
Moreover, the best result is achieved when using all five factors for debiasing
(\emph{event}, \emph{attribute}, \emph{context} are not used to create \texttt{K200-both-split}). 

\begin{figure}[t]
	\centering
	\captionsetup{font=small}
	\includegraphics[width=.8\columnwidth]{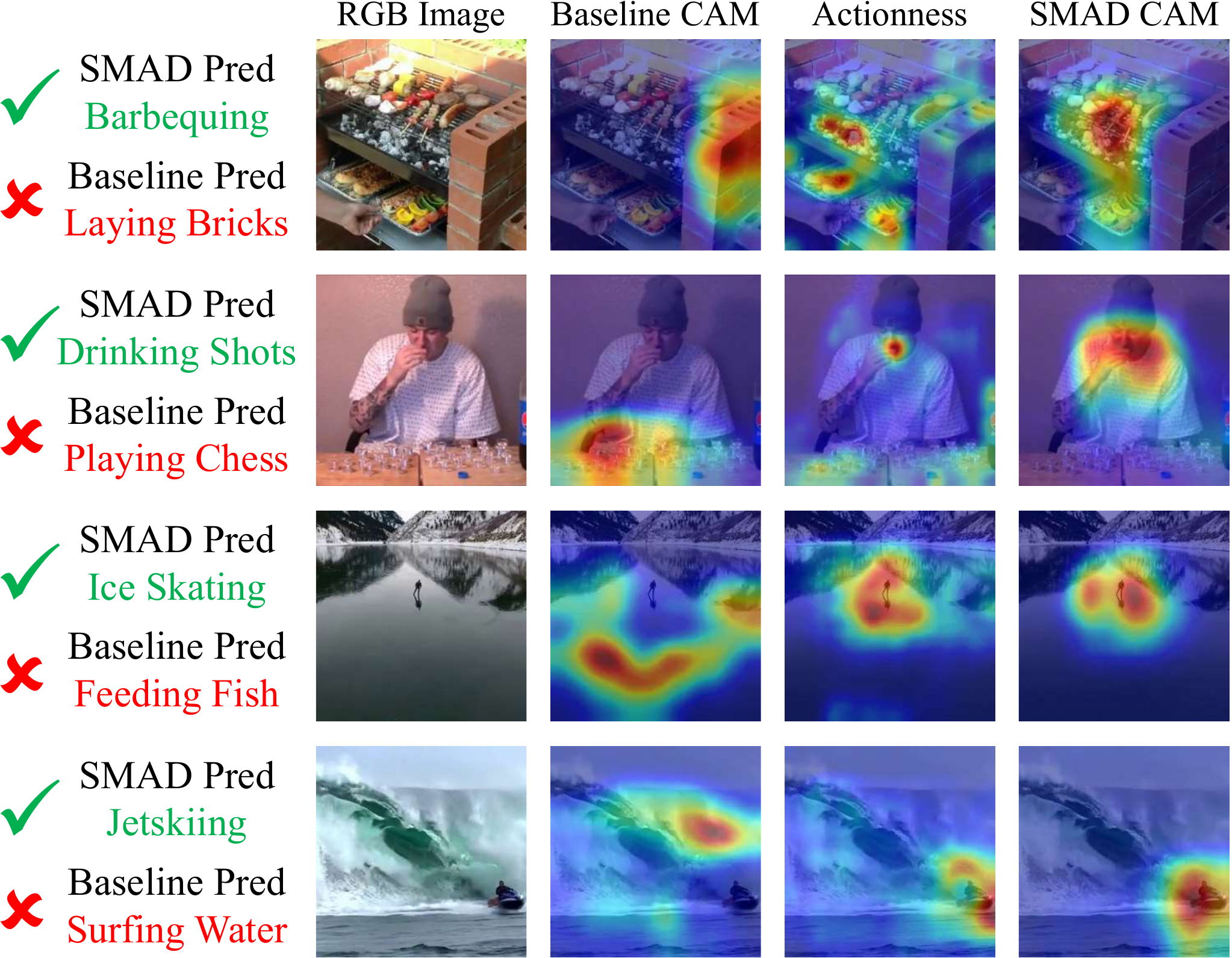}
	\vspace{-2mm}
	\caption{\textbf{The visualization of Actionness mask and CAM.\protect\footnotemark }}
	\label{fig-quali}
\end{figure}

\footnotetext{Visualization videos in \url{https://youtu.be/j1LA3y-UuEA}. }

\noindent\textbf{Qualitative Results.}
To qualitatively show how SAAR guides feature learning, 
we visualize the spatial-aware actionness mask predicted by SAAR
and the class activation maps (CAM)~\cite{zhou2016learning} of models 
trained with or without SMAD in Figure~\ref{fig-quali}.
Without debiasing, the rare scenes in action videos, 
\eg, brick grill, many chessman-like shots, transparent ice surface, and huge waves may mislead the model to give out wrong predictions.
With SMAD, models can learn to focus on human actions rather than scenes.

\subsection{Exploiting Web Media with OmniDebias}

\noindent \textbf{Web Data Help in Debiasing.}
To exploit the richness and diversity of web media, 
we propose joint training with both labeled datasets and unlabeled web datasets.
We first try to use the entire web dataset after teacher filtering for joint training, 
including both web image dataset \texttt{GG-all} and web video dataset \texttt{IG-all}.
Table~\ref{tab-dsdev2} shows that the performance improved by web media is considerable for $z$-unbiased test sets.
For both baselines, the gain on \texttt{ARAS-64} is around $10\%$.
The improvement on the normal test set \texttt{K200-val}, is milder ($ 1\sim4\% $) but also noticeable.

\noindent \textbf{Data Selection Strategy.}
Although web media contain novel $z \in \cZ$ that seldomly or never occurs in the original train set, 
the per action category $z$ distributions are still highly imbalanced.
To study the contribution of each portion of web media, 
we split each web dataset into 3 equal-sized parts: 
\texttt{bias}, \texttt{mid}, \texttt{unbias}.
We also randomly sample a third from the web dataset (\texttt{rand}) for comparison.
Using \texttt{bias} and \texttt{mid} leads to worse performance than \texttt{rand}.
Using \texttt{unbias}, however, not only surpasses other subsets, 
but also outperforms training with all web data.
With OmniDebias, the performance gap between $z$-unbiased test sets and \texttt{K200-val} is largely narrowed:
the gap shrinks by around 10\% Top1 for \texttt{ARAS-64} and around 5\% Top1 for \texttt{K200db-val}.
Besides re-distributed datasets, the improvement of OmniDebias can also be observed 
when trained on the full Kinetics dataset\footnote{We list the detailed results in the supplementary material. }.

\begin{table}[t]
	\centering
	\captionsetup{font=small}
	\resizebox{.8\linewidth}{!}{
		\tablestyle{10pt}{1.2}
	\begin{tabular}{c|cccc}
		\shline
		Model & Aux-Data & \texttt{ARAS-64}  & \texttt{K200db-val} & \texttt{K200-val} \\ 
		\shline
		\multirow{6}{*}{2D} & \texttt{None} 	& 55.8    & 53.7 	& 76.2 \\
		& \texttt{GG-bias}    & 60.0 \inc{4.2} 	& 53.9 \inc{0.2}   	& 76.0 \dec{0.2}   	\\
		& \texttt{GG-mid}     & 60.5 \inc{4.7}	& 55.1 \inc{1.4}   	& 76.3 \inc{0.1}   	\\
		& \texttt{GG-rand}    & 64.1 \inc{8.3}	& 57.5 \inc{3.8}   	& 77.2 \inc{1.0}  	\\
		& \texttt{GG-all}     & 65.5 \inc{9.7} 	& 58.4 \inc{4.7}   	& 77.8 \inc{1.6}   	\\
		& \texttt{GG-unbias}  & 68.3 \inc{12.5}	& 60.0 \inc{6.3}   	& 77.4 \inc{1.2}   	\\ 
		\shline
		\multirow{4}{*}{3D} & \texttt{None}		& 51.0    & 51.9 	& 75.1 	\\
		& \texttt{IG-rand}    & 62.3 \inc{11.3}	& 56.8 \inc{4.9}   	& 77.4 \inc{2.3}  	\\
		& \texttt{IG-all}     & 63.4 \inc{12.4}	& 58.5 \inc{6.6}   	& 78.3 \inc{3.2}  	\\
		& \texttt{IG-unbias}  & 64.7 \inc{13.7}	& 59.8 \inc{7.9}   	& 78.1 \inc{3.0}   	\\ 
		\shline
	\end{tabular}}
	\vspace{1mm}
	\caption{\textbf{OmniDebias.} 
		We jointly train \texttt{K200db-train} with different web dataset splits. 
		The improvement for $z$-unbiased test sets (\texttt{K200db-val}, \texttt{ARAS-64}) is much larger than \texttt{K200-val}.}
	\label{tab-dsdev2}
	\vspace{-4mm}
\end{table}

\section{Conclusion}
In this work, we seek to mitigate the \textit{generic} representation bias in action recognition.
We propose SMAD and OmniDebias: SMAD integrates multi-head adversarial training and spatial-aware feature reweighting for algorithm debiasing, 
while OmniDebias exploits the rich diversity of web data efficiently for dataset debiasing.
When combined, two components lead to excellent debiasing performance and 
perform far better on either artificially split test sets or manually collected out-of-distribution ones.

\appendix
\begin{table}[h]
	\centering
	\captionsetup{font=small}
	\resizebox{.8\linewidth}{!}{
	\tablestyle{10pt}{1.2}
	\begin{tabular}{c|ccc}
		\shline
		split factor   & dataset    &  $I(A, S)$ &  $I(A, O)$ \\ 
		\shline
		\multirow{2}{*}{scene}  & K200db-train & 0.466 \rinc{0.069}  & -   \\
		& K200db-val   & 0.374 \bdec{0.114}  & -  \\ 
		\shline
		\multirow{2}{*}{object} & K200db-train & - & 0.473 \rinc{0.033}   \\
		& K200db-val   & -     & 0.431 \bdec{0.104}   \\ 
		\shline
		\multirow{2}{*}{\begin{tabular}[c]{@{}c@{}}scene \& \\ object\end{tabular}} & K200db-train & 0.442 \rinc{0.045}  & 0.480 \rinc{0.040}   \\ 
		& K200db-val   & 0.393 \bdec{0.095}  & 0.456 \bdec{0.079}   \\ 
		\shline
	\end{tabular}}
	\vspace{2mm}
	\caption{
		\textbf{The NMI of action and different factors on different dataset splits. } 
		$A$, $S$, $O$ denote \emph{action}, \emph{scene} and \emph{object}. 
		\blue{Blue} and \red{red} marks denote the decrease or increase in NMI compared with random sampling (Larger NMI $\rightarrow$ more biased). }
	\label{tab:nmi_k200}
	\vspace{-11mm}
\end{table}

\section{Details on the Benchmark}

\subsection{How we construct \texttt{K200db-train/val}}

To build \texttt{K200db-train} and \texttt{K200db-val}, 
we first choose 200 categories in Kinetics-400 which have the most training samples.
When using \emph{scene} as the splitting factor\footnote{The pipeline can also be applied to other factors like \emph{object}. },
we first use a ResNet50 pretrained on Places365 to extract pseudo \emph{scene} labels for all videos.
Then, for each action class, we sort videos that belong to it by descending order of pseudo \emph{scene} label frequencies. 
For example, if 164 of 989 videos in the category `abseiling' belong to the \emph{scene} category `rope bridge', 
the pseudo \emph{scene} label frequency for each video will be $164 / 989 = 0.166$.
By doing so, videos with regular scenes are ranked on the top,
videos with rare scenes are ranked at the bottom. 
For each class, we choose at least\footnote{
	All videos of one \emph{scene} category should either belong to the training or testing set, 
	so the number might be larger than 50.} 
50 videos with the lowest pseudo \emph{scene} label frequency as testing videos, 
randomly sample 400 videos from the remaining videos as training videos. 
Finally, we construct \texttt{K200db-[train/val]} with 80000 and 10236 videos respectively.
We provide the training and testing sets file lists in supplementary materials and some sample videos in \url{https://youtu.be/j1LA3y-UuEA}.
We also list the normalized mutual information (NMI) of action and different factors in Table~\ref{tab:nmi_k200}. 
The re-distributed training set is more biased than random sampling (NMI \red{increases}),
while the re-distributed testing set is less biased than random sampling (NMI \blue{decreases}).

\subsection{How we construct \texttt{ARAS}}
For each action category in Kinetics-400, 
we try to collect action videos with scenes that rarely occur in Kinetics-400 from Youtube.
However, due to the taxonomy of Kinetics-400, such videos do not always exist for every category:
1. scenes might be limited by the category names (`biking through snow', \etc);
2. the action itself might restrict the scene (`jumping into pool', `snowmobiling', \etc).
For the applicable 104 action categories, 
we come up with one or several rare scenes after browsing videos in Kinetics-400 and 
then use the combinations of action and rare scenes as queries on Youtube. 
For example, we query `ice skating' with the scene `clear lake', query `eating watermelon' with the scene `farm'. 
After data collection, we conduct de-duplication to make sure that ARAS does not overlap with the Kinetics-400 training set.
We get around ten videos for each action category and trim one or several\footnote{Only if the untrimmed video is a collection of highlights.} clips from each video.
To construct a balanced test set, we only keep ten clips for each category.
\texttt{ARAS-104} contains 1,038 clips from 920 videos, while \texttt{ARAS-64} contains 640 clips from 579 videos.
We provide sample clips of \texttt{ARAS} in \url{https://youtu.be/j1LA3y-UuEA}.

\begin{table}[t]
	\centering
	\captionsetup{font=small}
	\resizebox{.7\linewidth}{!}{
		\tablestyle{10pt}{1.2}
		\begin{tabular}{ccc}
			\hline
			Dataset Split & \texttt{GoogleImage} & \texttt{InstagramVideo} \\ 
			\shline
			Raw & 6M & 1.1M \\ 
			\shline
			Positive-200 & 1.1M & 341K \\ 
			Positive-200 $1/3$ & 379K & 114K \\ 
			\shline
			Positive-400 & 1.9M & 481K \\
			Positive-400 $1/3$ & 620K & 160K \\ 
			\shline
	\end{tabular}}
	\vspace{1mm}
	\caption{\textbf{The sizes of the web datasets. }}
	\label{tab:size}
	\vspace{-4mm}
\end{table}

\begin{table}[t]
	\centering
	\captionsetup{font=small}
	\resizebox{.8\linewidth}{!}{
		\tablestyle{10pt}{1.2}
		\begin{tabular}{c|cccc}
			\shline
			$I(A, S)$ & \texttt{GG-200} & \texttt{IG-200} & \texttt{GG-400} & \texttt{IG-400}\\ 
			\shline 
			\texttt{all}    & 0.476  & 0.399  & 0.458  & 0.388  \\
			\texttt{rand}   & 0.481  & 0.412  & 0.464  & 0.402  \\
			\texttt{bias}   & 0.801  & 0.746  & 0.779  & 0.731  \\
			\texttt{mid}    & 0.687  & 0.625  & 0.665  & 0.604  \\
			\texttt{unbias} & 0.371  & 0.302  & 0.352  & 0.295  \\ 
			\shline
	\end{tabular}}
	\vspace{2mm}
	\caption{
		\textbf{The NMI of \textit{action} and \textit{scene} in different splits of web media datasets. } 
		$A$, $S$ denote \emph{action} and \emph{scene}. 
		The dataset size of \texttt{rand}, \texttt{bias}, \texttt{mid}, \texttt{unbias} are $1 / 3$  of \texttt{all}. }
	\label{tab:nmi_omni}
	\vspace{-8mm}
\end{table}

\section{OmniDebias Dataset Statistics}

We use \texttt{GoogleImg} and \texttt{InsVideo} in the large-scale web media dataset OmniSource~\cite{duan2020omni}\footnote{Dataset available at
	\url{https://github.com/open-mmlab/mmaction2/tree/master/tools/data/omnisource}} as auxiliary datasets for debiasing.
During pre-processing, we first use teacher networks 
(TSN for \texttt{GoogleImage}, SlowOnly for \texttt{InstagramVideo}) 
to filter out positive samples in the web dataset.
Then we sort web images/videos that belong to an action class 
by descending order of pseudo \emph{scene} label frequencies (as described in \textbf{How we construct \texttt{K200db-train/val}}).
Based on the pseudo \emph{scene} frequencies, we divide all positive examples into three equal-sized parts: 
\texttt{[web]-bias} (high frequency), \texttt{[web]-mid} (middle frequency), \texttt{[web]-unbias} (low frequency).
Only \texttt{[web]-unbias} is used for co-training by OmniDebias.
We list the size of each web dataset in Table~\ref{tab:size}.
\textbf{Raw} denotes the raw web datasets without teacher filtering of 400 action classes.  
\textbf{Positive-200/400} denotes the positive examples belong to 200 action classes in \texttt{K200db} or 400 action classes in Kinetics-400.
$\mathbf{1/3}$ denotes the size of each divided part, \eg, \texttt{unbias}. 
Table~\ref{tab:size} shows that the web datasets (Positive-[200/400], $1/3$) used for co-training by OmniDebias are of the same magnitude as \texttt{K200db-train} or \texttt{K400-train}.

In Table~\ref{tab:nmi_omni}, we calculate the NMI between \emph{action} and \emph{scene} on different splits of web datasets.
The NMI of the \texttt{[web]-unbias} split is much smaller than any other split or the entire web dataset,
which indicates that \texttt{[web]-unbias} is less biased to the factor \emph{scene}.

\begin{table}[t]
	\centering
	\captionsetup{font=small}
	\resizebox{.7\linewidth}{!}{
		\tablestyle{10pt}{1.2}
	\begin{tabular}{c|cccc}
		\hline
		Test Set  	& Baseline & SoftCE & KL & SoftCE + KL \\ \shline
		ARAS-64     & 51.0    & 56.3    & 56.6    & 58.4     \\ \hline
		K200-unbias & 51.9    & 54.7    & 54.5    & 55.0     \\ \hline
	\end{tabular}} 
	\vspace{1mm}
	\caption{\textbf{Ablation on loss functions for soft labels. }}
	\label{tab-lossfunc}
	\vspace{-5mm}
\end{table}

\section{Experiment Setting}
For experiments on Kinetics, we use an ImageNet-pretrained 3-segment TSN~\cite{wang2016temporal} with ResNet50 backbone as the 2D baseline and use a randomly initialized SlowOnly-8x8~\cite{feichtenhofer2019slowfast} with ResNet18 backbone as the 3D one.
Each experiment runs on the server with 8 1080Ti GPUs.
RandomResizedCrop is used for data augmentation during the training of all experiments.

For 2D experiments, the batch-size for each GPU is 32. 
The training lasts 100 epochs.
We use 0.01 as the initial learning rate and drop the learning rate to its $1 / 10$ at the end of the 40th epoch and the 80th epoch.
During testing, we uniformly sample 25 frames with CenterCrop as input.

For 3D experiments, the batch-size for each GPU is 16.
Following~\cite{feichtenhofer2019slowfast}, the training lasts 196 epochs.
We perform linear warmup during the first 34 epochs.
A half-period cosine schedule~\cite{loshchilov2016sgdr} is adopted for learning rate decay: the learning rate at the n-th epoch is  $\eta \cdot 0.5 [\mathbf{cos}(\frac{n}{n_{max}} \pi)  + 1]$, where $n_{max}$ is the maximum training epochs (196 here) and the base learning rate $\eta$ is set to 0.2. 
During testing, we uniformly sample 10 clips with CenterCrop as input.

\section{Experiment Results}

\subsection{Choice of adversarial losses.}
Both SoftCE and KLDiv loss can be used for soft-label.
In practice, we find that the best debiasing result is achieved when we combine them as the loss function, 
which achieves 0.3\% and 1.8\% Top-1 acc gain on \texttt{K200db-val} and \texttt{ARAS-64} than each individual loss (Table~\ref{tab-lossfunc}).

\begin{table}[t]
	\centering
	\captionsetup{font=small}
	\resizebox{.7\linewidth}{!}{
	\tablestyle{8pt}{1.2}
	\begin{tabular}{c|ccc}
		\hline
		Aux-Data           & ARAS-64 & K200db-val & K200-val \\ \shline
		\texttt{GG-rand}   & 58.9  & 54.1 & 75.0  \\ \hline
		\texttt{GG-all}    & 61.7  & 55.1 & 75.9  \\ \hline
		\texttt{GG-unbias} & 65.3  & 56.2 & 75.2  \\ \shline
		\texttt{[GG+IG]-rand}   & 65.6  & 58.0 & 78.2  \\ \hline
		\texttt{[GG+IG]-all}    & 64.7  & 58.7 & 78.3 \\ \hline
		\texttt{[GG+IG]-unbias} & 69.2  & 60.7 & 78.7  \\ \hline
	\end{tabular}}
	\vspace{1mm}
	\caption{Results for the combinations: \texttt{K200db-train} + \texttt{GG} and \texttt{K200db-train} + \texttt{[GG+IG]}. Our data selection strategies also work for these two settings.}
	\label{tab:tab1}
	\vspace{-5mm}
\end{table}

\begin{table}[t]
	\centering
	\captionsetup{font=small}
	\resizebox{.7\linewidth}{!}{
	\tablestyle{10pt}{1.2}
	\begin{tabular}{c|c|cc}
		\hline
		Model                 & Aux-Data           	& \texttt{ARAS-104}	& \texttt{K400-val} \\ \shline
		\multirow{2}{*}{2D}  & \texttt{GG-all}  & 58.3  & 71.7  \\ \cline{2-4}
		& \texttt{GG-unbias}    & 60.2 & 71.3     \\ \shline
		\multirow{2}{*}{3D} & \texttt{[GG+IG]-all}    & 58.3 		& 70.7 	\\ \cline{2-4}
		& \texttt{[GG+IG]-unbias} & 60.2     	& 71.0  \\ \hline
	\end{tabular}}
	\vspace{2mm}
	\caption{On real-world large-scale datasets, 
		auxiliary web dataset and the data selection strategy still considerably improves the performance of \texttt{ARAS-104}.}
	\label{tab:dsfull}
	\vspace{-5mm}
\end{table}

\subsection{Qualitative Results}
We provide the visualization videos of CAM~\cite{zhou2016learning} and 
learned actionness mask corresponding to Figure \red{7} in \url{https://youtu.be/j1LA3y-UuEA}. 
Besides that, we also provide several additional examples. 

\subsection{Web Data Help in Debiasing}
In main paper Table \red{2}, 
for the 3D baseline trained on \texttt{K200db-train}, 
we report the performance of OmniDebias using \texttt{[GG+IG]} as the auxiliary web dataset.
In main paper Table \red{8}, for the 3D baseline trained on \texttt{K200db-train}, 
we report the performance of OmniDebias using \texttt{IG} as the auxiliary web dataset 
to validate our data selection strategy.
In Table~\ref{tab:tab1}, 
we demonstrate the conclusion of main paper Table \red{8} also holds for the two combinations 
\texttt{K200db-train} + \texttt{GG} and \texttt{K200db-train} + \texttt{[GG+IG]} over the 3D baseline.
Besides re-distributed datasets, 
the improvement of OmniDebias can also be observed 
when trained on the full Kinetics dataset, 
as shown in Table~\ref{tab:dsfull}.

\bibliographystyle{splncs04}
\bibliography{egbib}
\end{document}